\documentclass[letterpaper, 10 pt, conference]{ieeeconf}

\IEEEoverridecommandlockouts                              

\usepackage{cite}
\usepackage{float}
\usepackage{amssymb,amsfonts}
\usepackage{amsmath}
\usepackage{algorithmic}
\usepackage{graphicx}
\usepackage{textcomp}
\usepackage{xcolor}
\usepackage{gensymb}
\usepackage{subcaption}
\usepackage{multirow}

\usepackage{amsthm}
\newtheoremstyle{mystyle}
  {}
  {}
  {\itshape}
  {}
  {\bfseries}
  {.}
  { }
  {}
\theoremstyle{mystyle}
\newtheorem{definition}{Definition}

\newtheorem{assumption}{Assumption}

\DeclareMathOperator*{\E}{\mathbf{E}}
\DeclareMathOperator*{\argmax}{arg\,max}

\newcommand{\modify}[1]{{#1}} 
\def\BibTeX{{\rm B\kern-.05em{\sc i\kern-.025em b}\kern-.08em
    T\kern-.1667em\lower.7ex\hbox{E}\kern-.125emX}}

\usepackage[font=normal]{caption}

\begin{document}

\title{
Learning Visual Affordances with Target-Orientated Deep Q-Network\\to Grasp Objects by Harnessing Environmental Fixtures
}
\author{Hengyue Liang, Xibai Lou, Yang Yang and Changhyun Choi
\thanks{*This work was in part supported by the MnDRIVE Initiative on Robotics, Sensors, and Advanced Manufacturing.}
\thanks{$\dagger$The authors are with the University of Minnesota, Minneapolis, MN 55455, USA. {\tt\small \{liang656, lou00015, yang5276, cchoi\}@umn.edu}}%
}

\maketitle

\begin{abstract}
This paper introduces a challenging object grasping task and proposes a self-supervised learning approach. The goal of the task is to grasp an object which is not feasible with a single parallel gripper, but only with harnessing environment fixtures (e.g., walls, furniture, heavy objects). This Slide-to-Wall grasping task assumes no prior knowledge except the partial observation of a target object. Hence the robot should learn an effective policy given a scene observation that may include the target object, environmental fixtures, and any other disturbing objects. 
We formulate the problem as visual affordances learning for which Target-Oriented Deep Q-Network (TO-DQN) is proposed to efficiently learn visual affordance maps (i.e., Q-maps) to guide robot actions. Since the training necessitates robot's exploration and collision with the fixtures, TO-DQN is first trained safely with a simulated robot manipulator and then applied to a real robot. We empirically show that TO-DQN can learn to solve the task in different environment settings in simulation and outperforms a standard and a variant of Deep Q-Network (DQN) in terms of training efficiency and robustness. The testing performance in both simulation and real-robot experiments shows that the policy trained by TO-DQN achieves comparable performance to humans. 
\end{abstract}


\section{Introduction}
\label{introduction}
\begin{figure}[t]
    \centering
    \includegraphics[width=0.99\linewidth, height=0.27\linewidth]{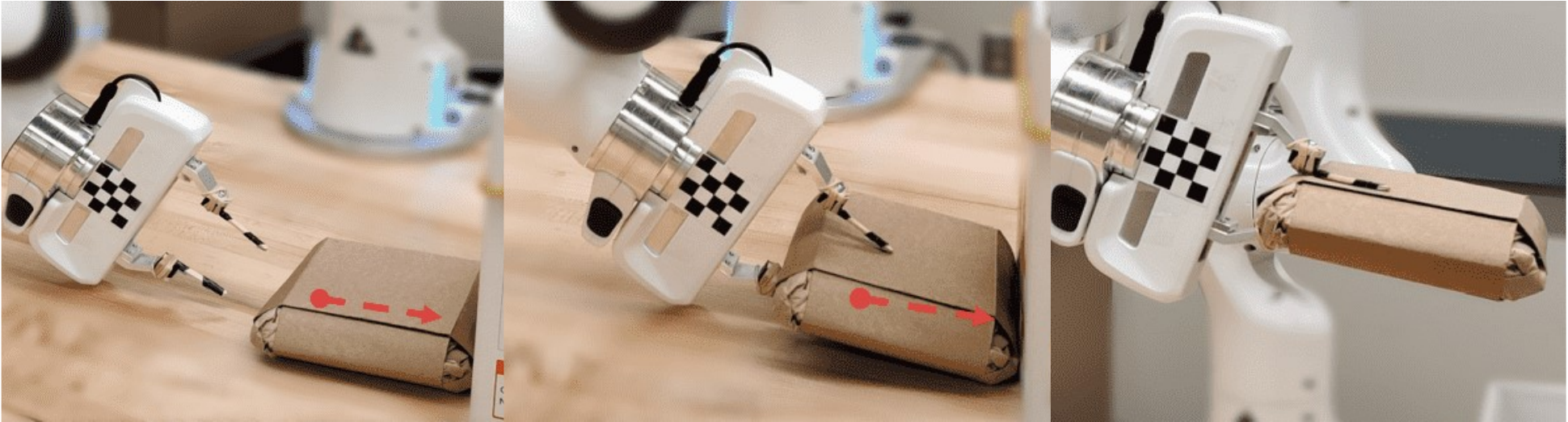}%
    \vspace{-3pt}
    \caption{\small{\textbf{A demo of the Slide-to-Wall grasping with Shovel-and-Grasp (SaG) action primitive.} SaG: the gripper 
    approaches to 
    the action point $(i, j)$ (red dot) along the action direction $\phi$ (red arrow). After reaching the action point, the gripper closes and lifts the object. If the target object is not close to the wall, the gripper is not able to lift the object and only pushes the object forward.}}%
    \label{fig:Sag Example}
\end{figure}

Imagine that you are about to grasp a flat object on a table. If you cannot grasp with one hand, you may use the other hand to constrain the object's motion. If a wall is accessible near the object, you may slide the object to the wall to facilitate grasping, as shown in Fig.~\ref{fig:Sag Example}. Such a task is named Slide-to-Wall grasping~\cite{eppner2015exploitation} that requires both sliding and lifting motions. 
Given an observation without assumptions of prior information about the task scene, it is challenging to perform the Slide-to-Wall grasping for a robot.
In order to succeed, the robot should 1) understand how the scene is composed of different objects, such as target objects, environmental fixtures, and any other non-target objects, 2) localize the target object, 3) determine where to slide the object, and 4) predict the consequences of the interactions between the robot, objects, and fixtures.
%

While robotic object grasping has been actively studied in the past few decades~\cite{miller2004graspit,levine2018learning,pinto2016supersizing,mahler2017dex,lou2020learning,mousavian2019graspnet}, object grasping utilizing environmental constraints has been less explored. 
\cite{eppner2015exploitation, eppner2015planning} studied various human grasping strategies to motivate robotic gripper designs and introduced the Slide-to-Edge and the Slide-to-Wall grasping strategies that enable robot hands to exploit an edge and a wall of a planar surface for grasping a flat and a cylindrical object, respectively. While pioneered grasping with environmental constraints, they employed a simple visual servoing to initiate grasping with a pre-defined control strategy in structured environments, limiting its generalization capability to even slightly different settings.  

In this paper, we propose a data-driven method to achieve the Slide-to-Wall grasping in unstructured environments, by formulating the problem as a Markov Decision Process (MDP).  
As the transition model (e.g., how the scene changes given an action) is unknown, we employ a model-free reinforcement learning (RL) algorithm, Deep Q-Network (DQN)~\cite{mnih2015human} to learn a grasping policy. Contrary to the standard DQN that searches the entire action space, we provide the target object information to DQN so that it learns Q-maps efficiently and generalizes to various objects. We name this variant Target-Oriented Deep Q-Network (TO-DQN). 

In addition to the need of numerous exploration samples for DQN to learn meaningful Q-values, learning contact-rich manipulation with environmental constraints is potentially harmful to physical robot arms. In order to practice safe exploration for the Slide-to-Wall grasping task, we train our network in simulation with a virtual robot manipulator and environment and evaluate it with a real robot. Our extensive evaluation in simulation and real world validates that our policy can generalize well in the real world and attain almost human-level performance.
The main contributions of this paper are as follows:
\begin{itemize}
\item{We introduce a challenging robotic manipulation task where the robot can only achieve successful grasping by exploiting certain environmental fixtures with a sequence of actions (see Sec.~\ref{Sec: Problem formulations}).
}
\item{We propose TO-DQN to train the visual affordance network (Q-network) using target object information, for which we devise a novel update rule conditioned on the target object region (see Sec.~\ref{subsec: kidqn}).
}
\item{We empirically show that TO-DQN outperforms a standard DQN and a variant of it in terms of sample efficiency, task performance, generalization capability with respect to environment's changes (see Sec.~\ref{Sec: Exp}).}
\end{itemize}

\section{Related Works}

\textbf{Visual Affordance Maps and Action Primitives}: 
Using Q-maps to guide action primitives has recently demonstrated the ability to solve complex, challenging manipulation tasks. \cite{zeng2018learning} studies clearing densely cluttered objects and \cite{yang2020deep, danielczuk2019mechanical} focus on searching and grasping target objects under occluded conditions. \cite{zakka2020form2fit} achieves efficient assembly by learning from disassembly, and \cite{wu2020spatial} guides navigation robots to push target objects to the desired positions. Though learning the Q-maps of action primitives sacrifices some motion flexibility compared to continuous control in the motor space, it provides an efficient way to build task-level manipulation policies. In this paper, we adopt this task-level control strategy with a hybrid action primitive. 

\textbf{Deep Q-Network}: DQN is a model-free RL alorithm that features a deep neural network as Q-function approximator, a discrete action space, and experience replay for stable learning convergence~\cite{mnih2015human}. It has achieved superhuman performance in playing Atari games~\cite{mnih2015human}, where a convolutional neural network (CNN) is trained end-to-end from raw image input to the output action space. Many works have improved the learning process (e.g., double Q-learning~\cite{hasselt2016deep} to mitigate the overestimation, prioritized experience replay~\cite{schaul2015prioritized} to improve the effective use of data, dueling network~\cite{wang2016dueling} to enable state-dependent actions, Rainbow DQN~\cite{hessel2017rainbow} combining  the aforementioned improvements into one). Our approach is based on DQN but designed for more challenging scenarios where 1) the action space is multi-dimensional rather than 1-dimensional and 2) the environment is not consistent across learning and testing (e.g., training in simulation and testing in real world). 

\textbf{Object Detection}: Modern object detection pipelines have an object proposal stage (e.g., bounding box~\cite{girshick2014rich,girshick2015fast,ren2015faster,redmon2016you} or object masks~\cite{pinheiro2015learning, pinheiro2016learning, dai2016instance, li2017fully}) followed by a classification stage.
Mask R-CNN~\cite{he2017mask} provides state-of-the-art performance for instance segmentation and detection problems with RGB images. 
In robotic manipulation, \cite{danielczuk2019segmenting} proposed a variant of Mask R-CNN trained with only synthetic depth images. We follow a similar idea to \cite{danielczuk2019segmenting} but use both grayscale and depth channels to achieve better object instance segmentation results (see Sec.~\ref{subsubsec: r-cnn}). We further leverage a Siamese network~\cite{bromley1994signature} trained with RGB-D images to detect the target object in a reference image. The combination of segmentation and Siamese networks provides a flexible way to detect target objects and to generalize to novel objects. 

\textbf{Manipulation Exploiting Environmental Constraints}: 
Considering environmental constraints for robotic manipulation appears in the early works~\cite{lozano1984automatic, mason1985mechanics}. For the grasping problems, \cite{kaneko2000scale} mimics different human strategies to model practical robot grasping actions, including utilizing the edge effect between the gripper and the supporting surface. \cite{eppner2015exploitation} further investigates on this problem and focuses on compliant gripper designs to exploit such constraints. There are also various studies targeting at robot motion planning to achieve grasping from such exploitation. For example, \cite{kappler2012templates} enable a humanoid robot to slide objects for a better grasp by planning with prior knowledge of object models.
\cite{maeda2001planning, salvietti2015modeling} study motions to use compliant fingers effectively to grasp objects with the help of the supporting floor. In all the above works, the environment is well controlled prior to robot operations and many prior knowledge was assumed to be known, while our approach relax such assumptions as delineated in Sec.~\ref{Sec: Problem formulations}.

\newcommand{\targetobj}{\ensuremath{\mathcal{O}}}
\newcommand{\fixture}{\ensuremath{\mathcal{F}}}
\newcommand{\disturbobj}{\ensuremath{\mathcal{D}}}
\newcommand{\obs}{\ensuremath{\mathcal{S}}} 
\newcommand{\actions}{\ensuremath{\mathcal{A}}}
\newcommand{\action}{\ensuremath{a}}
\newcommand{\state}{\ensuremath{s}}
\newcommand{\traj}{\ensuremath{\tau}}

\section{Problem Formulations and Preliminaries}
\label{Sec: Problem formulations}
\begin{figure}[t]
    \centering
    \includegraphics[height=0.12\textwidth, width=0.49\textwidth]{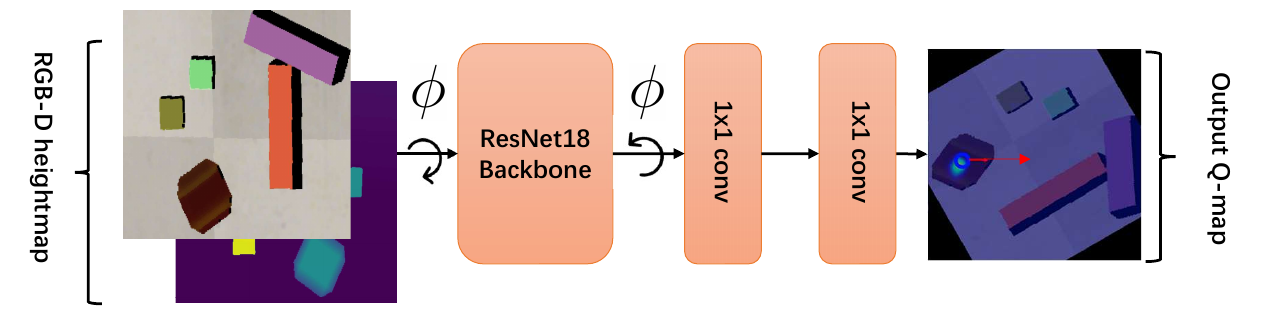}%
    \vspace{-3pt}
    \caption{\small{\textbf{The Q-network structure} approximating $Q(s, a^{i,j,\phi})$. The input to the network is a 4-channel RGB-D heightmap with pixel resolution $1~mm$ of the workspace. The input is first rotated by $\phi$ and sent into a ResNet18 backbone (substituting the first convolutional layer to match the 4-channel input). The intermediate features from the ResNet backbone are then rotated back by $-\phi$ and fed into a convolutional module to derive the final output $Q(s, a^{i,j,\phi};\theta)$, where $\theta$ is the network parameters.}}%
    \label{fig:model struct}
\end{figure}
In this section, we formally define the Slide-to-Wall grasping task as follows:
\begin{definition}
Given observations of workspace $\obs$, the \textbf{goal} of the Slide-to-Wall grasping task is to find a sequence of actions, $\traj = \langle \action_1, \action_2, \cdots, \action_N \rangle$ where $N$ is the length of the task episode, in a workspace to successfully grasp a target object $\targetobj$ by harnessing environmental fixtures $\fixture$.
\end{definition}
Our approach uses RGB-D images as the observation $\obs$, and the action $\action$ is determined from the $Q$-maps (Fig.~\ref{fig:model struct} and see Sec.~\ref{Subsec: vamap} for details). As the problem requires sequential decision making, we formulate a Markov Decision Process (MDP) as delineated in Sec.~\ref{Subsec: mdp}. We assume as follows:
\begin{assumption}
The target object $\targetobj$ is not graspable with one robot manipulator due to its geometric shape, and is only side-graspable with the interactions between the manipulator and the fixtures $\fixture$ in the workspace.
\end{assumption}
\begin{assumption}
The workspace images $\obs$ have partial observation of a target object $\targetobj$, one or more fixtures $\fixture$, and one or more disturbing objects $\disturbobj$.
\end{assumption}
\begin{assumption}
A reference image $I_o$ of the target object $\mathcal{O}$ is the only prior knowledge, and there is no visual and physical knowledge about fixtures $\mathcal{F}$ and disturbing objects $\mathcal{D}$.
\end{assumption}
Hence the robot should learn visual features of both fixtures $\fixture$ and disturbing objects $\disturbobj$ from its own experience. Due to the limited prior knowledge, we cannot foresee what happens to the workspace given an action $\action$ (i.e., transition probability is unknown). This motivates us to apply a target-oriented, model-free RL method to solve the task (see Sec.~\ref{subsec: kidqn}).

\subsection{Visual Affordance Maps}
\label{Subsec: vamap}
As the action space of DQNs is discrete~\cite{mnih2015human}, the action $\action$ needs to be defined in a discrete coordinate. 
In our approach, we define $\action \in \mathbb{N}^3$ by 2D action point coordinates $(i, j)$ and an action direction $\phi$ in degrees.
To be precise, we denote $a^{i,j,\phi}$ as the action $\action$ on $i$-th row and $j$-column of the discretized grid with $\phi$ action angle or direction.
The quality score of an action $a^{i,j,\phi}$ is denoted by $Q(s, a^{i, j, \phi}) \in \mathbb{R}^{W \times H}$, where $s \in \obs$ is an observation of the workspace and $W$ and $H$ are the width and height of the discretized workspace, respectively. Given a default angle $\phi_0$, the quality scores $Q(s, a^{i, j, \phi_0})$, $\forall i, j$, define a mapping to the physical robot workspace\footnote{In this paper, $Q(s, a^{i, j, \phi_0}) \in \mathbb{R}^{64 \times 64}$ is mapped to the square workspace size $50^2~cm^2$.}.
This mapping, named as \emph{visual affordance map (Q-map)}, provides the robot with the affordance measures of the action $\action$ in the current state $s$.

\subsection{Shovel-and-Grasp Action Primitive}
\label{Def: SaG primitve}

In the Slide-to-Wall grasping task, the action $\action$ represents a hybrid action primitive, Shovel-and-Grasp (SaG), as shown in Fig. \ref{fig:Sag Example}. It combines pushing and grasping such that SaG pushes the object if the object is far from any fixtures $\fixture$, or SaG puts one of the fingers underneath the object, tilts it, closes the fingers, and lifts it up if the object is close to a fixture $\fixture$. \modify{Note that even the gripper closes at the end of each SaG action, it may only achieve pushing behavior when a target object \targetobj~is far from \fixture}. As defined, SaG is only effective when action $\action$ is on the object. Otherwise, it either makes no changes of the workspace if applied on background or is hazardous due to collisions if applied on the fixtures.

\section{Learning of Visual Affordance Maps}
\label{Sec: training affordance}
%
To learn the Q-maps $Q(\state, \action^{i, j, \phi})$ effectively and reliably, we formulate the problem as an MDP and propose to solve it through TO-DQN.
\subsection{Markov Decision Process}
\label{Subsec: mdp}
An MDP is defined as a tuple $( \mathcal{S}, \mathcal{A}, \mathcal{P}, \mathcal{R}, \rho_0 )$, where $\mathcal{S}$ is the set of environment states, $\mathcal{A}$ is a finite set of actions, $\mathcal{P}$ is the transition probabilities, $\mathcal{R}$ is the reward function and $\rho_0$ is the distribution of the initial states. The goal is to find a policy $\pi^* = \argmax_{\pi}J(\pi)$ that maximizes the expected total reward:
\begin{equation}
    J(\pi) := \E_{\traj \sim \pi}[\sum_{t=0}^{T}\gamma^{t}r_t]
\end{equation}
where $\traj$ denotes the trajectory following the policy $\pi$, $r_t$ is the received reward at each step, and $\gamma \in [0, 1)$ is the discount factor balancing the short and long term rewards and ensuring mathematical convergence~\cite{white2001markov}.
\begin{figure*}[t]
\centering
\begin{subfigure}{.245\textwidth}
  \centering
  \includegraphics[height=0.82\linewidth, width=0.95\linewidth]{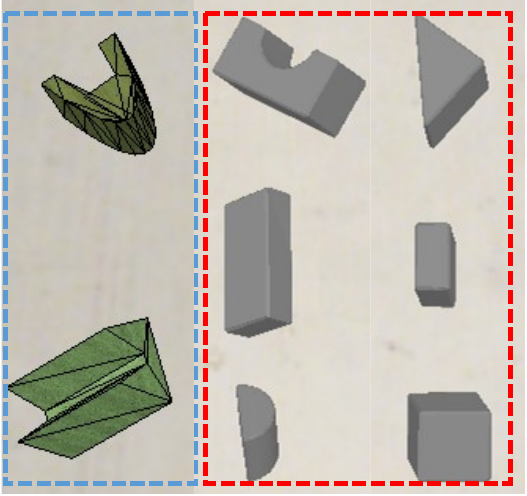}%
  \vspace{-2pt}
  \caption{\small{Disturbing Objects}}
  \label{Subfig: disturbs}
\end{subfigure}%
\begin{subfigure}{.245\textwidth}
  \centering
  \includegraphics[height=0.82\linewidth, width=0.95\linewidth]{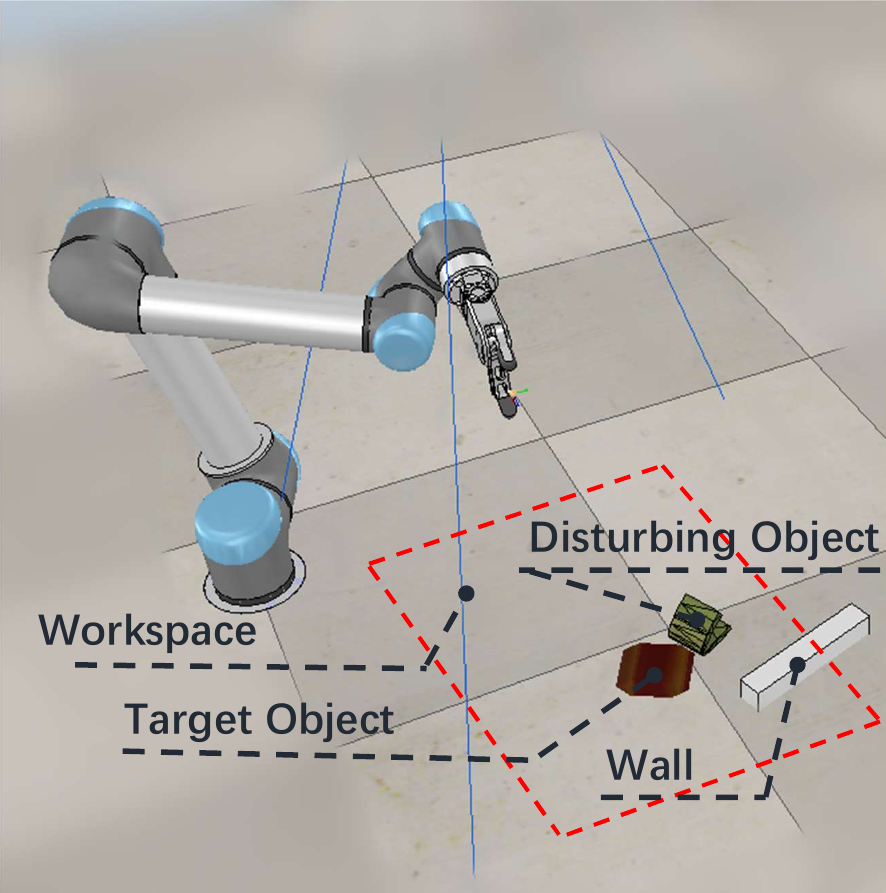}%
  \vspace{-2pt}
  \caption{\small{Training scene (Sim)}}
  \label{Fig: train scene}
\end{subfigure}%
\begin{subfigure}{.245\textwidth}
  \centering
  \includegraphics[height=0.82\linewidth, width=0.95\linewidth]{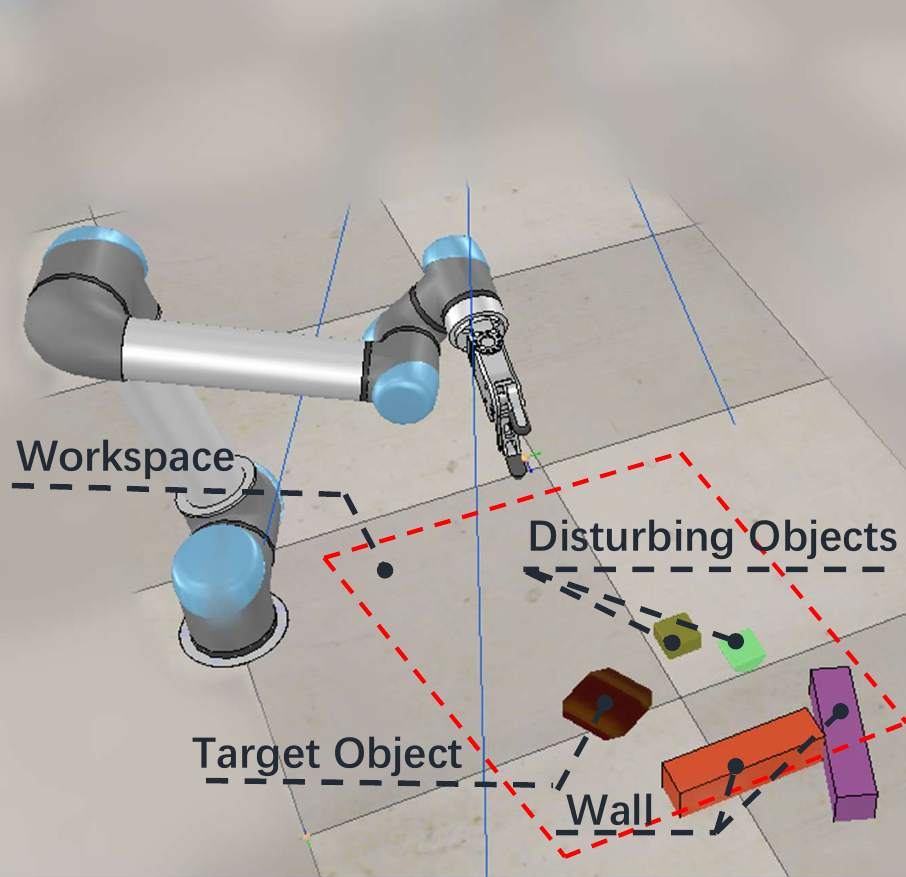}%
  \vspace{-2pt}
  \caption{\small{Testing scene (Sim)}}
  \label{Fig: test scene}
\end{subfigure}
\begin{subfigure}{.245\textwidth}
  \centering
  \includegraphics[height=0.82\linewidth, width=0.95\linewidth]{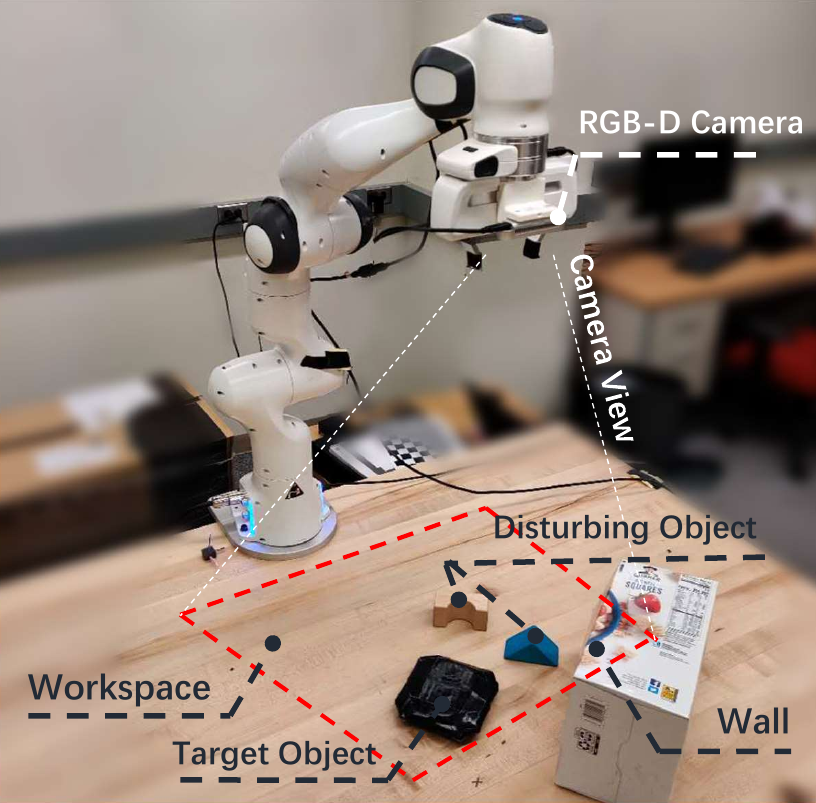}%
  \vspace{-2pt}
  \caption{\small{Real scene}}
  \label{Fig: Real Scene}
\end{subfigure}%
\caption{\small{\textbf{Experiment scene examples}.  
\textbf{(a) Disturbing objects} used in simulation during training (blue box) and testing (red box).
\textbf{(b) Training scenes} in simulation include a target object, one static wall with random position and orientation close to one corner of the workspace, one randomly dropped disturbing object.
\textbf{(c) Testing scenes} in simulation include a target object, one or two static walls, one or two randomly dropped disturbing objects.
\textbf{(d) Real-robot scene} An RGB-D camera is mounted on the Panda robot end-effector. The camera observes the workspace and captures an RGB-D heightmap at the fixed home position. In the workspace, we place a target object (flat box), other random toy blocks as the disturbing objects, and a large box as the wall.}}%
\label{fig: scene examples}
\end{figure*}

\subsection{Deep Q-Network}
\label{Subsec: dqn}
Because of the MDP formulation, learning the Q-maps can be reduced to learning the action-value functions (Q-functions) by DQN \cite{sutton1998introduction,mnih2015human}, which is defined as:
\begin{equation}
    Q^{\pi}(s,a;\theta) := \mathbf{E}_{s,a,\pi}[\sum^{\infty}_{t=1}\gamma^t r_t]
    \label{Action value function}
\end{equation}
where $s \in \obs$ and $a \in \actions$ represent the observation and action respectively, $\pi$ denotes the policy, and $\theta$ represents the parameters of a deep neural network approximating the quality score $Q(\state,\action)$ function.

For the Slide-to-Wall task, the transition probabilities $\mathcal{P}$ is unknown as the task involves contact-rich dynamics and unknown physical properties (e.g., friction, weights, center of mass).
Therefore, we treat the transitions as stochastic and DQN improves $Q$ function approximation by updating $\theta$ at each training step:
\begin{equation}
    \theta_{t+1} \leftarrow \theta_{t} - \alpha(y_t^Q - Q(s_t, a_t; \theta_t)) \nabla_{\theta} Q (s_t, a_t; \theta) 
\label{DQN Update Eq}
\end{equation}
where $\alpha$ is the learning rate, $y_t^Q$ is the update target value as
\begin{equation}
    y_t^Q = 
    \begin{cases}
    r_t ~, ~ \text{if $s_t$ is a terminal state} \\
    r_t + \gamma \max_{a} Q(s_{t+1}, a; \theta^{-}) ~ , ~ \text{otherwise}
    \end{cases}
\end{equation}
where $\theta^{-} = \theta^{t_0}$ for some $t_0 < t$ as a fixed value and updated by $\theta^{-} = \theta_t$ once every predefined step to ensure network stability during training \cite{mnih2015human}.  
By substituting $a_t$ with $a_t^{i, j, \phi}$, \eqref{DQN Update Eq} naturally turns into the DQN update to approximate $Q(s, a^{i, j, \phi})$.

\subsection{Target-Oriented DQN (TO-DQN)}
\label{subsec: kidqn}

As mentioned in Sec. \ref{Def: SaG primitve}, SaG is only meaningful when executed at a point on the target object. We notice that the object knowledge facilitates the learning of Q-maps, and hence propose the following target-oriented Q-function as:
\begin{equation}
    Q(s,a^{i, j, \phi}) := \begin{cases}
    0, ~ \text{if $a^{i, j, \phi} \notin \mathcal{R}_{s, a}^\phi$}  \\
    \mathbf{E}_{s,a,\pi}[\sum^{\infty}_{t=1}\gamma^t r_t], ~ \text{otherwise}
    \end{cases}
    \label{Eq. TO-DQN}
\end{equation}
where $\mathcal{R}_{s, a}^\phi$ is the effective action region pertinent to the object $\targetobj$. The values on the ineffective areas is strictly $0$ rather than any other values computed by the future rewards. To get valid \eqref{Eq. TO-DQN} approximations, we modify the DQN update \eqref{DQN Update Eq} accordingly by inducing the knowledge $\mathcal{R}_{s, a}^\phi$:
\begin{equation}
\begin{aligned}
    \theta_{t+1} \leftarrow \theta_{t} - \alpha \sum_{a^{i, j, \phi} \in \mathcal{A}}  & (y_t^{{i, j, \phi}} - Q(s_t, a^{i, j, \phi}; \theta_t)) \cdot \\ & \nabla_{\theta} Q (s_t, a^{i, j, \phi}; \theta) 
\end{aligned}
\label{TO-DQN Update Eq}
\end{equation}
where $y_t^{{i, j, \phi}}$ is the new target value defined as:
\begin{equation}
y_t^{{i, j, \phi}} = 
\begin{cases}
0, ~ \text{if}~\text{$a^{i, j, \phi} \notin \mathcal{R}_{s, a}^\phi$} \\
Q(s_t, a^{i, j, \phi}; \theta_t),  ~ \text{if}~\text{$a^{i, j, \phi} \in \mathcal{R}_{s, a}^\phi$ and $a^{i, j, \phi} \neq a_t$} \\
\hat{y}, ~ \text{otherwise}
\end{cases}
\label{TO-DQN Update Target Eq}
\end{equation}
and 
\begin{equation}
    \hat{y} = \begin{cases}
    r_t ~, ~ \text{if $s_t$ is a terminal state} \\
    r_t + \gamma \max_{a} Q(s_{t+1}, a^{i, j, \phi}; \theta^{-}) ~ , ~ \text{otherwise.}
    \end{cases}
\end{equation}
The well-trained Q-maps, $Q(s,a^{i, j, \phi}; \theta)$, $\forall i, j, \phi$, will thus be able to guide an effective policy $\pi$ by suppressing actions on the irrelevant regions in the workspace. 

\subsection{Target Region Detection}
\label{Subsec: Region}
To obtain the target object knowledge $\mathcal{R}_{s, a}^\phi$, we employ a two-stage, sequential detection method combining an instance segmentation network and an object identification network. 
We use a variant of Mask R-CNN \cite{he2017mask} to propose object instance segmentation masks. Then, a Siamese network \cite{bromley1994signature} trained with task-relevant objects is used to find the correct target mask by comparing each segmented object image with the target object image $I_o$.
Note that this detection method does not assume a fixed number of object classes and generalizes to novel objects as long as an observation $I_o$ of the target object $\targetobj$ is available.


\section{Experiment Settings and Results}
\label{Sec: Exp}
In this section, we empirically show the benefits of TO-DQN with the considered manipulation task. In the experiments, an RGB-D camera captures the heightmaps of the square workspace at a static viewpoint before the robot executes an SaG action at each step. All neural network models are built with PyTorch 1.5.1. \cite{paszke2019pytorch}, optimized using the Stochastic Gradient Decent (SGD) with learning rate $1 \times 10^{-5}$, momentum $0.9$, and weight-decay $2 \times 10^{-5}$. The simulation experiments are built through CoppeliaSim \cite{rohmer2013v} 4.0.0 with bullet engine 2.83. In the real-robot scenes, Intel RealSense D415 camera is used as the visual sensor.

The networks to detect the target object is trained separately prior to the SaG experiment. We train the visual affordance network (Q-network) for picking SaG actions in simulated manipulation scenes, as shown in Fig. \ref{Fig: train scene}, and test the trained network in different simulation settings (Fig.~\ref{Fig: test scene}) and the real-world scenes (Fig.~\ref{Fig: Real Scene}) to evaluate how our approach handles the environmental changes and the reality gaps. From this experiment, we aim to investigate:
\begin{itemize}
    \item Does the target region detection reliably find the object region?
    \item Can TO-DQN solve the Slide-to-Wall grasping via an effective sequence of actions?
    \item Is TO-DQN able to show reliable performance in unseen settings?
    \item Does TO-DQN outperform other baselines?
\end{itemize}

\subsection{Detecting the Target Object Region}
\subsubsection{Object Instance Segmentation}
\label{subsubsec: r-cnn}
\begin{figure}[t]
    \vspace{4pt}
    \centering
    \includegraphics[width=0.95\linewidth, height=0.3\linewidth]{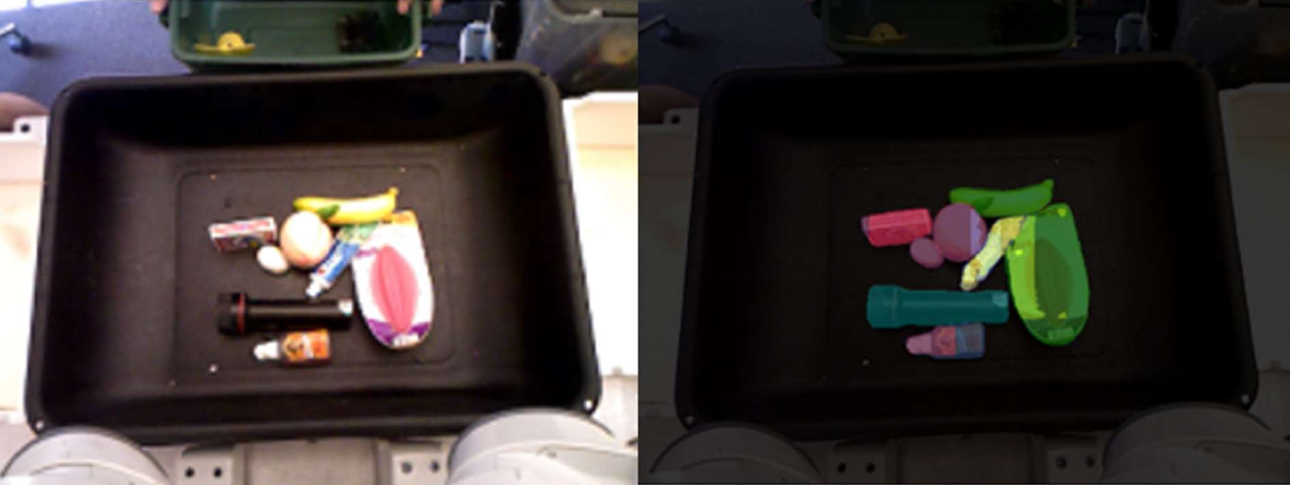}%
    \caption{\small{\textbf{An example of the object instance segmentation} in Sec. \ref{subsubsec: r-cnn}. The picture is from the WISDOM dataset~\cite{danielczuk2019segmenting}.}}%
    \label{fig:mask_visual}
\end{figure}

We use a variant of Mask R-CNN and train the network with the low-res real-world images from the WISDOM dataset~\cite{danielczuk2019segmenting}.
Our network consists of a pre-trained ResNet50 backbone on the CoCo dataset \cite{lin2014microsoft} and a modified two-class classification (background and foreground objects). The input channels to our model are `gray, depth, depth (GDD)' respectively instead of `depth, depth, depth (DDD)' as in SD Mask-RCNN\cite{danielczuk2019segmenting}. The reason for this choice is two-fold:
\begin{itemize}
    \item The edge information in grayscale images is the crucial part for object segmentation. 
    \item The reality gap of grayscale images is relatively lower than that of the RGB images, which has been shown in many robotic manipulation studies \cite{mahler2017dex}\cite{gualtieri2018learning}\cite{detry2017task}. 
\end{itemize}
To compare with SD Mask R-CNN, we 
pick a subset of $50$ images to train and the remaining $250$ images are reserved for evaluation, following the data split in~\cite{danielczuk2019segmenting}.
The evaluation result of the object segmentation is reported in TABLE~\ref{Table: sd mask rcnn} with the standard Average Precision (AP) and Average Recall (AR) criteria used in CoCo benchmarks~\cite{lin2014microsoft}. 
SD Mask R-CNN is trained with synthetic depth images only~\cite{danielczuk2019segmenting}. Fine-tuned Mask R-CNN methods are Mask R-CNN models fine-tuned with either RGB, depth, and gray images~\cite{danielczuk2019segmenting}.
According to TABLE~\ref{Table: sd mask rcnn}, our method outperforms the compared methods by a large margin. While \cite{danielczuk2019segmenting} \modify{focused on experimenting Sim-to-Real segmentation using a large amount of synthetic depth data, our work demonstrates that using a small number of real RGB-D annotated data can achieve much better performance.} 
A visualization of the mask generation by our method is shown in Fig.~\ref{fig:mask_visual}.

\begin{table}[t]
\caption{\small{Average Precision (AP) and Average Recall (AR) of the object instance segmentation methods. (see Sec.~\ref{subsubsec: r-cnn} for details)
}}%
\begin{tabular}{c|c|c}
\hline
Method & ~~ AP $(0.50: 0.95)$ ~~ & AR ($100$)\\ 
\hline
SD Mask R-CNN \cite{danielczuk2019segmenting}   & $0.356$ & $0.465$ \\ 
\hline
Fine-tuned Mask R-CNN (color) & $0.385$ & $0.613$ \\
\hline
Fine-tuned Mask R-CNN (depth) & $0.331$ & $0.546$ \\
\hline
Fine-tuned Mask R-CNN (gray) & $0.680$ & $0.746$ \\
\hline
Ours & $\textbf{0.763}$ & $\textbf{0.796}$ \\
\hline
\end{tabular}%
\label{Table: sd mask rcnn}
\end{table}
\subsubsection{Target Object Identification}
\label{subsec: Target Object Identification}
To identify the target object with an object image $I_o$, we use the Siamese network~\cite{bromley1994signature} with RGB-D inputs.
To train this network, we collected $100$ RGB-D images patches each of randomly placed target objects $\targetobj$, disturbing objects $\disturbobj$, and wall fixtures $\fixture$, before the robot starts the Slide-to-Wall manipulation tasks.  
\modify{The network is trained with triplet loss~\cite{chechik2010large} using the anchor images ($I_o$) and sampled positive ($\targetobj$) and negative ($\disturbobj$, $\fixture$) examples from the collected images.} The Siamese network consists of three convolutional layers, followed by one flatten layer and two fully connected layers, and has input dimension $\mathbb{R}^{4 \times 120 \times 120}$ and output dimension $\mathbb{R}^{20}$.
As the target objects $\targetobj$, disturbing objects $\disturbobj$, and wall fixtures $\fixture$ are distinct in shape and appearance (See Fig. \ref{fig: scene examples}), the trained Siamese network can find the correct target object without error in all environments. 
%

%
\begin{figure}[t]
    \centering
    \includegraphics[width=1.0\linewidth]{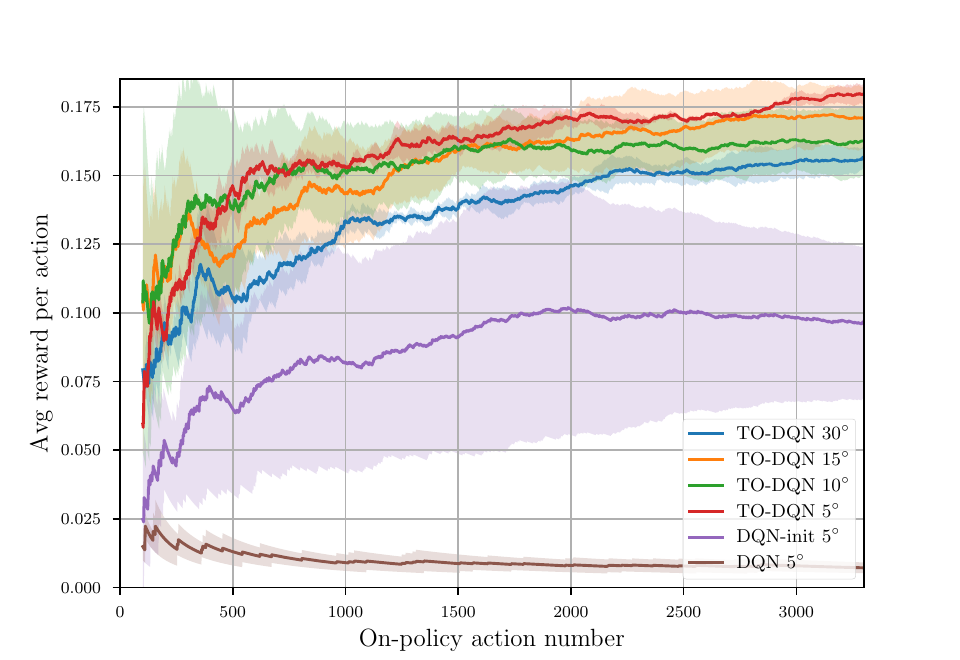}%
    \vspace{-2pt}
    \caption{\small{\textbf{Method-wise on-policy action efficiency during training.} The angle notation in each method denotes the SaG angle discretization interval of $\phi$ used during the training process.}}%
    \label{fig: training curve}
\end{figure}

\subsection{Slide-to-Wall Task Experiment Settings}
\label{SaG problem}
We train the Q-network in various simulation scenes. In each scene, we randomly place and color one target object, one disturbing object, and one static wall, as shown in Fig. \ref{Fig: train scene}. 
The target object during training is a $8 ~cm \times 8 ~ cm \times 3 ~ cm$ box. The wall height ranges from $8$ to $12 ~ cm$, length from $8$ to $12 ~ cm$, and $5 ~ cm$ thickness. The disturbing object in each training episode is picked from one of the blocks shown in Fig. \ref{Subfig: disturbs}.

During training, a reward $r_t = 1$ is assigned at the end of each episode if the grasp is successful; all other cases the robot receives $r_t = 0$. \modify{The knowledge $\mathcal{R}^{\phi}_{s, a}$ in \eqref{TO-DQN Update Target Eq} is derived by segmenting all object instances in the workspace as described in Sec.~\ref{subsubsec: r-cnn} and by comparing all segmented image patches with the target object image $I_o$~ as detailed in Sec.~\ref{subsec: Target Object Identification}.} 

We prepare several baselines: 
\textbf{Random} is the policy where SaG is executed at each step with a random angle $\phi$ and a random point on the target object region. It serves as a weak policy baseline but with perfect target object region. 
\textbf{Human} is the policy where a human user specifies the SaG point and angle during testing. This serves as a nearly optimal policy.
\textbf{DQN} is the standard DQN as explained in Sec.~\ref{Subsec: dqn}. \textbf{DQN-init} is the method where the Q-network is initially updated by TO-DQN \modify{\eqref{TO-DQN Update Eq}} within the first $200$ iterations to achieve low-value initialization on the region outside $\mathcal{R}_{s, a}^\phi$. Then the network follows standard DQN update.
\textbf{TO-DQN} is our method described in Sec.~\ref{subsec: kidqn}.
\begin{figure*}[t]
\centering
\begin{subfigure}{.115\textwidth}
  \centering
  \includegraphics[height=2.0\linewidth]{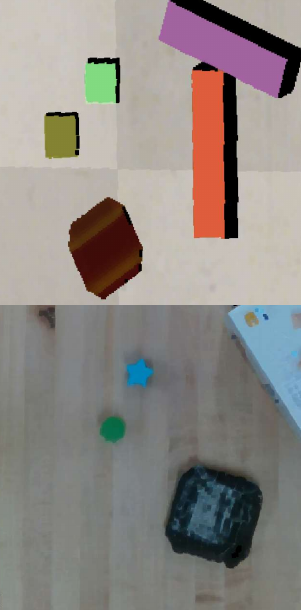}%
  \caption{\footnotesize\textbf{RGB}}
  \label{fig:Input Demo Color}
\end{subfigure}
\begin{subfigure}{.115\textwidth}
  \centering
  \includegraphics[height=2.0\linewidth]{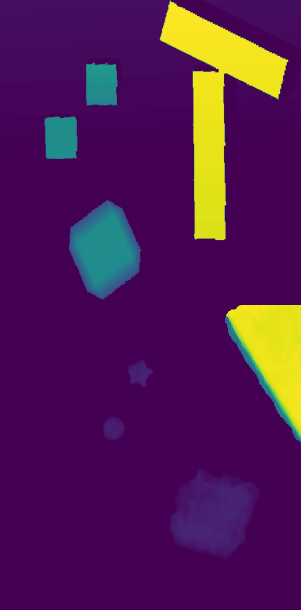}%
  \caption{\footnotesize\textbf{Depth}}
  \label{fig:Input Demo Depth}
\end{subfigure}
\begin{subfigure}{.115\textwidth}
  \centering
  \includegraphics[height=2.0\linewidth]{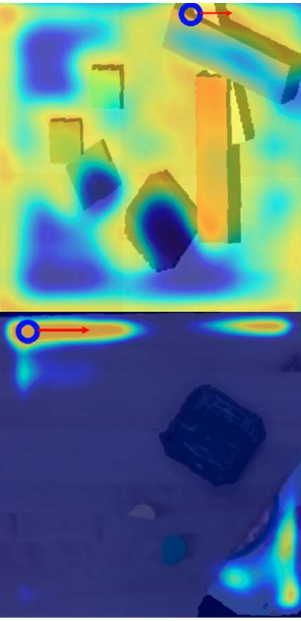}%
  \caption{\footnotesize\textbf{DQN}$^R$}
  \label{fig:q_map:dqn_raw}
\end{subfigure}
\begin{subfigure}{.115\textwidth}
  \centering
  \includegraphics[height=2.0\linewidth]{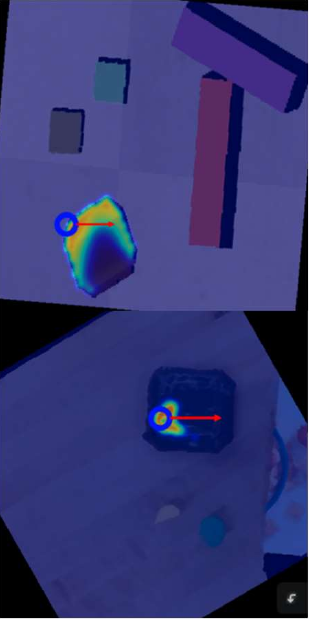}%
  \caption{\footnotesize\textbf{DQN}$^M$}
\end{subfigure}
\begin{subfigure}{.115\textwidth}
  \centering
  \includegraphics[height=2.0\linewidth]{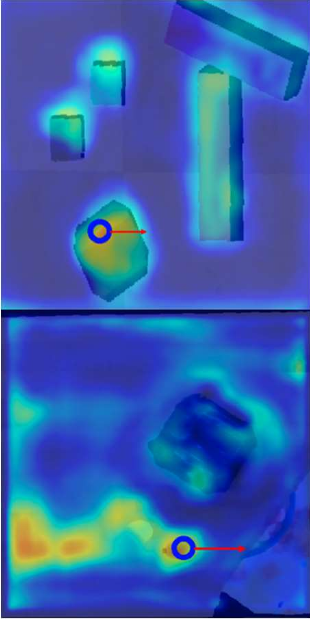}%
  \caption{\footnotesize\textbf{DQN-init}$^R$}
\end{subfigure}
\begin{subfigure}{.115\textwidth}
  \centering
  \includegraphics[height=2.0\linewidth]{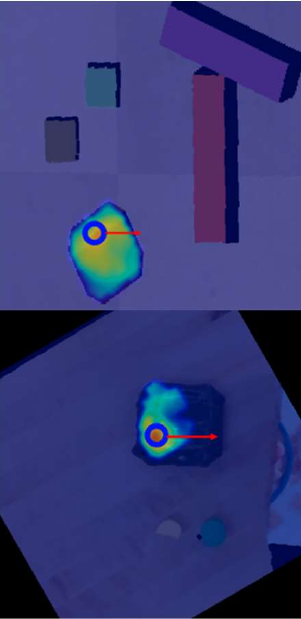}%
  \caption{\footnotesize\textbf{DQN-init}$^M$}
\end{subfigure}
\begin{subfigure}{.115\textwidth}
  \centering
  \includegraphics[height=2.0\linewidth]{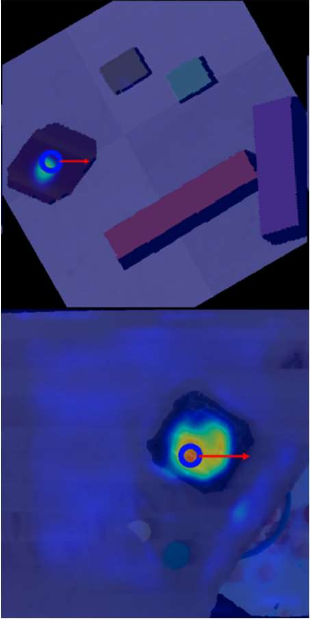}%
  \caption{\footnotesize\textbf{TO-DQN}$^R$}
\end{subfigure}
\begin{subfigure}{.115\textwidth}
  \centering
  \includegraphics[height=2.0\linewidth]{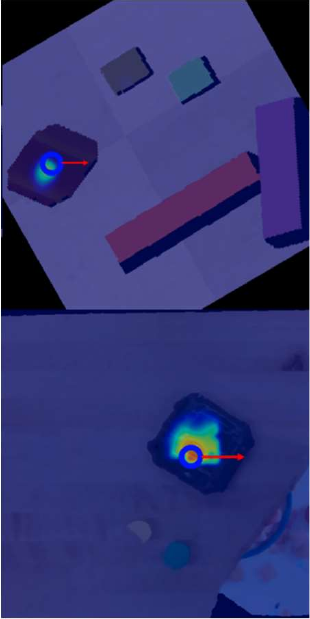}%
  \caption{\footnotesize\textbf{TO-DQN}$^M$}
  \label{fig:q_map:todqn_masked}
\end{subfigure}
\vspace{-2pt}
\caption{\small{\textbf{Examples of the color (a) and depth (b) input heightmaps} and \textbf{visualized affordance predictions (c)-(h)} in simulation (the top row) and real-robot testing (the bottom row). The blue circle and red arrow indicates the action point $(i, j)$ and direction $\phi$ provided by a greed policy on each affordance map. The $R$ and $M$ superscripts denote `raw' and `masked', respectively.}}%
\label{fig:Q_visual_real}
\end{figure*}

\subsection{Visual Affordance Network Training}
\label{Subsec: Visual Affordance Training}
\subsubsection{Training Settings}
We train the Q-network with $\epsilon$-greedy policy and discount factor $\gamma = 0.9$ during the robot on-policy self-exploration. The maximum number of actions per episode is set as $10$. When the robot successfully grasps the target object or the episode reaches the maximum number of actions, the environment will reset with a different scene setting. At each action step, we run one SGD update on the newest collected transition. Experience replay starts to carry out every step with $4$ randomly sampled terminal transitions and $4$ other transitions when the total action number reaches $200$.
\begin{figure}[t]
    \centering
    \includegraphics[width=1.0\linewidth]{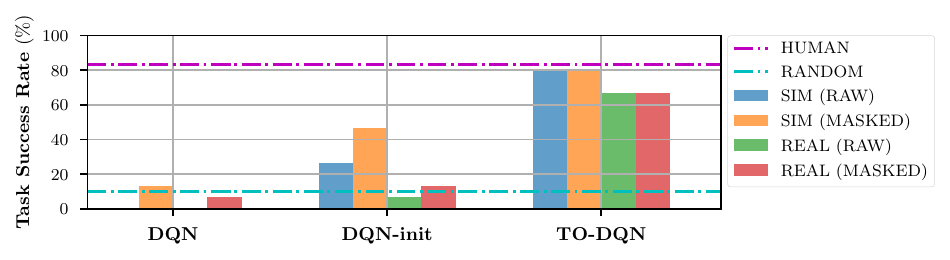}%
    \vspace{-4pt}
    \caption{\small{\textbf{The testing performance} of the trained affordance in simualtion and real-robot environments. `RAW': a greedy policy on the Q-maps, `MASKED': a greey policy on the Q-maps masked with the target object region $\mathcal{R}_{s, a}^\phi$.}}%
    \label{fig: testing stats}
\end{figure}
%
%
\subsubsection{Training Efficiency}
We first evaluate the training efficiency of the DQN-based methods. During the exploration phase, the Q-maps are filtered by $\mathcal{R}_{s, a}^\phi$ before applying $\epsilon$-greedy policy for all methods.

Fig.~\ref{fig: training curve} shows the average reward per action each method receives during training. 
To evaluate how the discretization intervals of the rotation angle $\phi$ affect the performance, we run with $\phi = 5^{\circ}, 10^{\circ}, 15^{\circ}, 30^{\circ}$. 
While \textbf{TO-DQN}s follow the similar learning curves, the \textbf{TO-DQN} $5^\circ$ curve shows the highest average reward with narrower variances. 
Note that the number of network parameters for all affordance networks is the same regardless of the discretization interval of $\phi$, as the visual encoder is agnostic to the rotation $\phi$ (Fig.~\ref{fig:model struct}). Thus smaller discretization intervals do not sacrifice the training efficiency in terms of training steps but provide more fine-grained actions to achieve better performance. 

Comparing the different training methods with the same interval $\phi$ (i.e., $\phi = 5^{\circ}$), \textbf{TO-DQN} outperforms the other two methods by a large margin. Although \textbf{DQN-init} achieves better performance than \textbf{DQN} in terms of training efficiency, the large gap between \textbf{DQN-init} and \textbf{TO-DQN} indicates that inducing the target object knowledge $\mathcal{R}_{s, a}^\phi$ benefits the learning throughout the training process, not only in the early stage. As indicated by Fig. \ref{fig: training curve}, \textbf{DQN} has not formed an effective policy within $3500$ training steps.
\subsection{Generalization Performance}
We also test the performance of the Q-networks both in simulation (as Fig. \ref{Fig: test scene}) and real-world experiments (as Fig. \ref{Fig: Real Scene}). 
In simulation, test scenes are generated with a varying number of walls ($1$ or $2$), different wall heights and width (from $3$ to $15 ~ cm$), and unseen disturbing objects as in Fig.~\ref{fig: scene examples}. The Q-networks trained with $\phi = 5^\circ$  angle discretization are tested. 
We add one additional failure criterion during testing, from which the episode ends if the robot tries to act on the wall fixtures.

We ran 15 tests on each method and the results are shown in Fig.~\ref{fig: testing stats}. Methods denoted by `RAW' represent the action is picked by a greedy policy on the entire visual affordance maps, while `MASKED' represents that the affordance predictions are filtered by $\mathcal{R}_{s, a}^\phi$ before applying the greedy policy.

\textbf{Human} does not achieve $100\%$ accuracy because of several unstable grasps caused by the uncertain contacts between the target object and the wall. In simulation testing, \textbf{TO-DQN} and \textbf{DQN-init} achieve task success rates higher than that of \textbf{Random}, indicating that effective projections from the input images to the output affordance maps have been established, and \textbf{TO-DQN} has derived a far better policy to handle the different settings. \textbf{DQN} is discouraging in this testing and \textbf{DQN (masked)} at best manages to achieve comparable performance to \textbf{Random} baseline. It implies that the policies from \textbf{DQN} and \textbf{DQN (masked)} cannot handle the environmental changes, which is consistent with the result from Fig.~\ref{fig: training curve}.

We observe the performance reduction for each method from simulation to real-robot experiment. The performance degradation is significantly larger in \textbf{DQN-init} than \textbf{TO-DQN}. 
To understand better, we further visualize the input and output of the visual affordance networks.
Fig. \ref{fig:Input Demo Color} and \ref{fig:Input Demo Depth} show the observation differences between the simulation and real-world environment (color and depth maps, respectively). 
The affordance predictions of each method are visualized in Fig.~\ref{fig:q_map:dqn_raw}-\ref{fig:q_map:todqn_masked}. Though \textbf{DQN-init} is able to demonstrate consistent behavior in simulation with and without $\mathcal{R}_{s, a}^\phi$ masking, the affordance maps are more disturbed in the real-world settings and cannot keep a robust and consistent prediction.
Only \textbf{TO-DQN} keeps consistent and effective predictions in both settings, showing more robustness to the unseen environment changes.

\section{Discussion and Future work}
In this paper, we proposed a challenging Slide-to-Wall manipulation task where successful grasps can only be achieved by proper interactions with the fixtures in the environment. To cope with potential changes in the environment settings and solve the tasks effectively, we proposed a vision-based learning method to utilize visual affordance maps and a side-grasp action primitive. We empirically showed that the proposed method, TO-DQN, can efficiently train a policy to complete the task for various configurations in simulation and achieves comparable performance to humans. In addition, the trained policy is less sensitive to the environment changes compared to other baselines, and can adapt well to real world scenarios directly even without fine-tuning. 

\bibliographystyle{./bibliography/IEEEtran}
\bibliography{./bibliography/IEEEabrv,./bibliography/IEEEexample}

\begin{thebibliography}{10}
\providecommand{\url}[1]{#1}
\csname url@samestyle\endcsname
\providecommand{\newblock}{\relax}
\providecommand{\bibinfo}[2]{#2}
\providecommand{\BIBentrySTDinterwordspacing}{\spaceskip=0pt\relax}
\providecommand{\BIBentryALTinterwordstretchfactor}{4}
\providecommand{\BIBentryALTinterwordspacing}{\spaceskip=\fontdimen2\font plus
\BIBentryALTinterwordstretchfactor\fontdimen3\font minus
  \fontdimen4\font\relax}
\providecommand{\BIBforeignlanguage}[2]{{%
\expandafter\ifx\csname l@#1\endcsname\relax
\typeout{** WARNING: IEEEtran.bst: No hyphenation pattern has been}%
\typeout{** loaded for the language `#1'. Using the pattern for}%
\typeout{** the default language instead.}%
\else
\language=\csname l@#1\endcsname
\fi
#2}}
\providecommand{\BIBdecl}{\relax}
\BIBdecl

\bibitem{eppner2015exploitation}
C.~Eppner, R.~Deimel, J.~Alvarez-Ruiz, M.~Maertens, and O.~Brock,
  ``Exploitation of environmental constraints in human and robotic grasping,''
  \emph{The International Journal of Robotics Research}, vol.~34, no.~7, pp.
  1021--1038, 2015.

\bibitem{miller2004graspit}
A.~Miller and P.~Allen, ``Graspit! a versatile simulator for robotic
  grasping,'' \emph{IEEE Robotics \& Automation Magazine}, vol.~11, no.~4, pp.
  110--122, 2004.

\bibitem{levine2018learning}
S.~Levine, P.~Pastor, A.~Krizhevsky, J.~Ibarz, and D.~Quillen, ``Learning
  hand-eye coordination for robotic grasping with deep learning and large-scale
  data collection,'' \emph{The International Journal of Robotics Research},
  vol.~37, no. 4-5, pp. 421--436, 2018.

\bibitem{pinto2016supersizing}
L.~Pinto and A.~Gupta, ``Supersizing self-supervision: Learning to grasp from
  50k tries and 700 robot hours,'' in \emph{2016 IEEE international conference
  on robotics and automation (ICRA)}.\hskip 1em plus 0.5em minus 0.4em\relax
  IEEE, 2016, pp. 3406--3413.

\bibitem{mahler2017dex}
J.~Mahler, J.~Liang, S.~Niyaz, M.~Laskey, R.~Doan, X.~Liu, J.~A. Ojea, and
  K.~Goldberg, ``Dex-net 2.0: Deep learning to plan robust grasps with
  synthetic point clouds and analytic grasp metrics,'' \emph{Robotics: Science
  and Systems (RSS)}, 2017.

\bibitem{lou2020learning}
X.~Lou, Y.~Yang, and C.~Choi, ``Learning to generate 6-dof grasp poses with
  reachability awareness,'' in \emph{2020 IEEE International Conference on
  Robotics and Automation (ICRA)}.\hskip 1em plus 0.5em minus 0.4em\relax IEEE,
  2020, pp. 1532--1538.

\bibitem{mousavian2019graspnet}
A.~Mousavian, C.~Eppner, and D.~Fox, ``6-dof graspnet: Variational grasp
  generation for object manipulation,'' in \emph{International Conference on
  Computer Vision (ICCV)}, 2019.

\bibitem{eppner2015planning}
C.~Eppner and O.~Brock, ``Planning grasp strategies that exploit environmental
  constraints,'' in \emph{2015 IEEE International Conference on Robotics and
  Automation (ICRA)}.\hskip 1em plus 0.5em minus 0.4em\relax IEEE, 2015, pp.
  4947--4952.

\bibitem{mnih2015human}
V.~Mnih, K.~Kavukcuoglu, D.~Silver, A.~A. Rusu, J.~Veness, M.~G. Bellemare,
  A.~Graves, M.~Riedmiller, A.~K. Fidjeland, G.~Ostrovski \emph{et~al.},
  ``Human-level control through deep reinforcement learning,'' \emph{Nature},
  vol. 518, no. 7540, p. 529, 2015.

\bibitem{zeng2018learning}
A.~Zeng, S.~Song, S.~Welker, J.~Lee, A.~Rodriguez, and T.~Funkhouser,
  ``Learning synergies between pushing and grasping with self-supervised deep
  reinforcement learning,'' in \emph{2018 IEEE/RSJ International Conference on
  Intelligent Robots and Systems (IROS)}.\hskip 1em plus 0.5em minus
  0.4em\relax IEEE, 2018, pp. 4238--4245.

\bibitem{yang2020deep}
Y.~Yang, H.~Liang, and C.~Choi, ``A deep learning approach to grasping the
  invisible,'' \emph{IEEE Robotics and Automation Letters}, 2020.

\bibitem{danielczuk2019mechanical}
M.~Danielczuk, A.~Kurenkov, A.~Balakrishna, M.~Matl, D.~Wang,
  R.~Mart{\'\i}n-Mart{\'\i}n, A.~Garg, S.~Savarese, and K.~Goldberg,
  ``Mechanical search: Multi-step retrieval of a target object occluded by
  clutter,'' in \emph{2019 International Conference on Robotics and Automation
  (ICRA)}.\hskip 1em plus 0.5em minus 0.4em\relax IEEE, 2019, pp. 1614--1621.

\bibitem{zakka2020form2fit}
K.~Zakka, A.~Zeng, J.~Lee, and S.~Song, ``Form2fit: Learning shape priors for
  generalizable assembly from disassembly,'' in \emph{2020 IEEE International
  Conference on Robotics and Automation (ICRA)}.\hskip 1em plus 0.5em minus
  0.4em\relax IEEE, 2020, pp. 9404--9410.

\bibitem{wu2020spatial}
J.~Wu, X.~Sun, A.~Zeng, S.~Song, J.~Lee, S.~Rusinkiewicz, and T.~Funkhouser,
  ``Spatial action maps for mobile manipulation,'' in \emph{Proceedings of
  Robotics: Science and Systems (RSS)}, 2020.

\bibitem{hasselt2016deep}
H.~v. Hasselt, A.~Guez, and D.~Silver, ``Deep reinforcement learning with
  double q-learning,'' in \emph{Proceedings of the Thirtieth AAAI Conference on
  Artificial Intelligence}, 2016, pp. 2094--2100.

\bibitem{schaul2015prioritized}
T.~Schaul, J.~Quan, I.~Antonoglou, and D.~Silver, ``Prioritized experience
  replay,'' \emph{arXiv preprint arXiv:1511.05952}, 2015.

\bibitem{wang2016dueling}
Z.~Wang, T.~Schaul, M.~Hessel, H.~Hasselt, M.~Lanctot, and N.~Freitas,
  ``Dueling network architectures for deep reinforcement learning,'' in
  \emph{International conference on machine learning}, 2016, pp. 1995--2003.

\bibitem{hessel2017rainbow}
M.~Hessel, J.~Modayil, H.~Van~Hasselt, T.~Schaul, G.~Ostrovski, W.~Dabney,
  D.~Horgan, B.~Piot, M.~Azar, and D.~Silver, ``Rainbow: Combining improvements
  in deep reinforcement learning,'' \emph{arXiv preprint arXiv:1710.02298},
  2017.

\bibitem{girshick2014rich}
R.~Girshick, J.~Donahue, T.~Darrell, and J.~Malik, ``Rich feature hierarchies
  for accurate object detection and semantic segmentation,'' in
  \emph{Proceedings of the IEEE conference on computer vision and pattern
  recognition}, 2014, pp. 580--587.

\bibitem{girshick2015fast}
R.~Girshick, ``Fast r-cnn,'' in \emph{Proceedings of the IEEE international
  conference on computer vision}, 2015, pp. 1440--1448.

\bibitem{ren2015faster}
S.~Ren, K.~He, R.~Girshick, and J.~Sun, ``Faster r-cnn: Towards real-time
  object detection with region proposal networks,'' in \emph{Advances in neural
  information processing systems}, 2015, pp. 91--99.

\bibitem{redmon2016you}
J.~Redmon, S.~Divvala, R.~Girshick, and A.~Farhadi, ``You only look once:
  Unified, real-time object detection,'' in \emph{Proceedings of the IEEE
  conference on computer vision and pattern recognition}, 2016, pp. 779--788.

\bibitem{pinheiro2015learning}
P.~O. Pinheiro, R.~Collobert, and P.~Doll{\'a}r, ``Learning to segment object
  candidates,'' in \emph{Advances in Neural Information Processing Systems},
  2015, pp. 1990--1998.

\bibitem{pinheiro2016learning}
P.~O. Pinheiro, T.-Y. Lin, R.~Collobert, and P.~Doll{\'a}r, ``Learning to
  refine object segments,'' in \emph{European conference on computer
  vision}.\hskip 1em plus 0.5em minus 0.4em\relax Springer, 2016, pp. 75--91.

\bibitem{dai2016instance}
J.~Dai, K.~He, Y.~Li, S.~Ren, and J.~Sun, ``Instance-sensitive fully
  convolutional networks,'' in \emph{European Conference on Computer
  Vision}.\hskip 1em plus 0.5em minus 0.4em\relax Springer, 2016, pp. 534--549.

\bibitem{li2017fully}
Y.~Li, H.~Qi, J.~Dai, X.~Ji, and Y.~Wei, ``Fully convolutional instance-aware
  semantic segmentation,'' in \emph{Proceedings of the IEEE Conference on
  Computer Vision and Pattern Recognition}, 2017, pp. 2359--2367.

\bibitem{he2017mask}
K.~He, G.~Gkioxari, P.~Doll{\'a}r, and R.~Girshick, ``Mask r-cnn,'' in
  \emph{Proceedings of the IEEE international conference on computer vision},
  2017, pp. 2961--2969.

\bibitem{danielczuk2019segmenting}
M.~Danielczuk, M.~Matl, S.~Gupta, A.~Li, A.~Lee, J.~Mahler, and K.~Goldberg,
  ``Segmenting unknown 3d objects from real depth images using mask r-cnn
  trained on synthetic data,'' in \emph{2019 International Conference on
  Robotics and Automation (ICRA)}.\hskip 1em plus 0.5em minus 0.4em\relax IEEE,
  2019, pp. 7283--7290.

\bibitem{bromley1994signature}
J.~Bromley, I.~Guyon, Y.~LeCun, E.~S{\"a}ckinger, and R.~Shah, ``Signature
  verification using a" siamese" time delay neural network,'' in \emph{Advances
  in neural information processing systems}, 1994, pp. 737--744.

\bibitem{lozano1984automatic}
T.~Lozano-Perez, M.~T. Mason, and R.~H. Taylor, ``Automatic synthesis of
  fine-motion strategies for robots,'' \emph{The International Journal of
  Robotics Research}, vol.~3, no.~1, pp. 3--24, 1984.

\bibitem{mason1985mechanics}
M.~Mason, ``The mechanics of manipulation,'' in \emph{Proceedings. 1985 IEEE
  International Conference on Robotics and Automation}, vol.~2.\hskip 1em plus
  0.5em minus 0.4em\relax IEEE, 1985, pp. 544--548.

\bibitem{kaneko2000scale}
M.~Kaneko, T.~Shirai, and T.~Tsuji, ``Scale-dependent grasp,'' \emph{IEEE
  Transactions on Systems, Man, and Cybernetics-Part A: Systems and Humans},
  vol.~30, no.~6, pp. 806--816, 2000.

\bibitem{kappler2012templates}
D.~Kappler, L.~Y. Chang, N.~S. Pollard, T.~Asfour, and R.~Dillmann, ``Templates
  for pre-grasp sliding interactions,'' \emph{Robotics and Autonomous Systems},
  vol.~60, no.~3, pp. 411--423, 2012.

\bibitem{maeda2001planning}
Y.~Maeda, H.~Kijimoto, Y.~Aiyama, and T.~Arai, ``Planning of graspless
  manipulation by multiple robot fingers,'' in \emph{Proceedings 2001 ICRA.
  IEEE International Conference on Robotics and Automation (Cat. No.
  01CH37164)}, vol.~3.\hskip 1em plus 0.5em minus 0.4em\relax IEEE, 2001, pp.
  2474--2479.

\bibitem{salvietti2015modeling}
G.~Salvietti, M.~Malvezzi, G.~Gioioso, and D.~Prattichizzo, ``Modeling
  compliant grasps exploiting environmental constraints,'' in \emph{2015 IEEE
  International Conference on Robotics and Automation (ICRA)}.\hskip 1em plus
  0.5em minus 0.4em\relax IEEE, 2015, pp. 4941--4946.

\bibitem{white2001markov}
C.~White, \emph{Markov decision processes}.\hskip 1em plus 0.5em minus
  0.4em\relax Springer, 2001.

\bibitem{sutton1998introduction}
R.~S. Sutton, A.~G. Barto \emph{et~al.}, \emph{Introduction to reinforcement
  learning}.\hskip 1em plus 0.5em minus 0.4em\relax MIT press Cambridge, 1998,
  vol. 135.

\bibitem{paszke2019pytorch}
A.~Paszke, S.~Gross, F.~Massa, A.~Lerer, J.~Bradbury, G.~Chanan, T.~Killeen,
  Z.~Lin, N.~Gimelshein, L.~Antiga \emph{et~al.}, ``Pytorch: An imperative
  style, high-performance deep learning library,'' in \emph{Advances in neural
  information processing systems}, 2019, pp. 8026--8037.

\bibitem{rohmer2013v}
E.~Rohmer, S.~P. Singh, and M.~Freese, ``V-rep: A versatile and scalable robot
  simulation framework,'' in \emph{2013 IEEE/RSJ International Conference on
  Intelligent Robots and Systems}.\hskip 1em plus 0.5em minus 0.4em\relax IEEE,
  2013, pp. 1321--1326.

\bibitem{lin2014microsoft}
T.-Y. Lin, M.~Maire, S.~Belongie, J.~Hays, P.~Perona, D.~Ramanan,
  P.~Doll{\'a}r, and C.~L. Zitnick, ``Microsoft coco: Common objects in
  context,'' in \emph{European conference on computer vision}.\hskip 1em plus
  0.5em minus 0.4em\relax Springer, 2014, pp. 740--755.

\bibitem{gualtieri2018learning}
M.~Gualtieri and R.~Platt, ``Learning 6-dof grasping and pick-place using
  attention focus,'' \emph{arXiv preprint arXiv:1806.06134}, 2018.

\bibitem{detry2017task}
R.~Detry, J.~Papon, and L.~Matthies, ``Task-oriented grasping with semantic and
  geometric scene understanding,'' in \emph{2017 IEEE/RSJ International
  Conference on Intelligent Robots and Systems (IROS)}.\hskip 1em plus 0.5em
  minus 0.4em\relax IEEE, 2017, pp. 3266--3273.

\bibitem{chechik2010large}
G.~Chechik, V.~Sharma, U.~Shalit, and S.~Bengio, ``Large scale online learning
  of image similarity through ranking,'' 2010.

\end{thebibliography}

\vspace{12pt}
\end{document}